**Flood of Techniques and Drought of Theories: Emotion Mining in Disasters**

Soheil Shapouri[1], Saber Soleymani[2], Saed Rezayi[2]

[1]Lehigh University, Department of Psychology, Bethlehem, PA, USA

[2]University of Georgia, School of Computing, Athens, GA, USA

**Author Note**

Soheil Shapouri https://orcid.org/0000-0001-9561-9751

Correspondence concerning this article should be addressed to Soheil Shapouri, Dept. of Psychology, 17 Memorial Dr. E, Bethlehem, PA, 18015. E-mail: sos523@lehigh.edu

Soheil Shapouri: https://scholar.google.com/citations?user=qzdp2SwAAAAJ&hl=en&oi=ao
Saber Soleymani: https://scholar.google.com/citations?user=ErHPjEUAAAAJ&hl=en&oi=ao
Saed Rezayi: https://scholar.google.com/citations?user=D_wiAi4AAAAJ&hl=en&oi=ao

**Abstract**

Emotion mining has become a crucial tool for understanding human emotions during disasters, leveraging the extensive data generated on social media platforms. This paper aims to summarize existing research on emotion mining within disaster contexts, highlighting both significant discoveries and persistent issues. On the one hand, emotion mining techniques have achieved acceptable accuracy enabling applications such as rapid damage assessment and mental health surveillance. On the other hand, with many studies adopting data-driven approaches, several methodological issues remain. These include arbitrary emotion classification, ignoring biases inherent in data collection from social media, such as the overrepresentation of individuals from higher socioeconomic status on Twitter, and the lack of application of theoretical frameworks like cross-cultural comparisons. These problems can be summarized as a notable lack of theory-driven research and ignoring insights from social and behavioral sciences. This paper underscores the need for interdisciplinary collaboration between computer scientists and social scientists to develop more robust and theoretically grounded approaches in emotion mining. By addressing these gaps, we aim to enhance the effectiveness and reliability of emotion mining methodologies, ultimately contributing to improved disaster preparedness, response, and recovery.

**Flood of Techniques and Drought of Theories: Emotion Mining in Disasters**

"Indian heatwave kills dozens over summer; nearly 25,000 fall ill," reports Reuters [70]. This headline is just one example of a growing global crisis: natural disasters and extreme weather events are escalating, with researchers predicting more frequent heatwaves, higher wind speeds of tropical cyclones, and more intense droughts in the near future [80]. Additionally, the advent of new technologies in the modern world has brought about unprecedented technological disasters such as nuclear accidents and oil spills [79]. To effectively combat these disasters, investigating affective responses to catastrophes is crucial as emotions significantly impact behavior and cognition and understanding human emotions during disasters can inform decision-making for disaster relief [82].

Psychologists and neuroscientists remain the primary investigators of emotions, but their traditional methods fail to capture human emotions in a timely manner. Human subject research with surveys requires institutional review board (IRB) approval but this is a time-consuming process [26] which hinders data collection during or shortly after disasters. Lab-based neuroimaging studies lack ecological validity; their results cannot be easily generalized to real-life situations [76]. So, emotion mining in social media led by computer scientists remains the most common tool for understanding human emotions expressed and communicated during disasters.

The past decade has witnessed an ever-growing sophistication of tools and techniques of emotion mining in the context of disasters. Manual detection of disaster-related tweets has been replaced by machine learning algorithms [84, 86]; disaster- and crisis-related datasets [54] and lexicons [4] have been created; and simple neural networks for extracting emotions have been substituted by deep learning approaches [67]. Despite these advances, several methodological issues, specifically ignoring insights from social and behavioral sciences prevail and persist which prevent the field from reaching its full potential. In this manuscript we first briefly review the literature on emotion mining in the context of disasters, then address methodological shortcomings of previous studies. We aim to provide a short review and a concise methodological discussion that can facilitate interdisciplinary collaboration between computer scientists interested in cutting-edge emotion mining techniques and psychologists concerned with rigor of methods and empirical support for theories.

**Floods of techniques; Emotion mining for understanding human responses to disasters**

Although sentiment analysis was around as early as the 90s (e.g., [87, 22]) and the terms sentiment analysis [62] and opinion mining [14] were coined more than two decades ago, it was the usefulness of text mining during 2010 Haiti earthquake [23] that sparked an interest in emotion mining during disasters. Sentiment analysis has a broad range of problem space ranging from topic modeling to emotion mining [45] but considering the importance of emotions in emergency situations [53] understating emotions by sentiment analysis has become an independent, important field of research. The main goal of this line of inquiry is to enhance situational awareness for better mitigation response [10] and the field has experienced advancements in different domains such as data collection, scope of the literature, emotion mining techniques, and applications.

***Platforms, data collection, and datasets***. Although researchers have used informational and content-based websites like Wikipedia for emotion mining in disasters (e.g., [25]), this field, like other domains of sentiment analysis, has focused largely on social media. Among different social media networks Twitter has been always the most popular option for emotion mining [72] and this holds true for disasters as well. Although there are studies conducted on Facebook [34], Instagram [33], TikTok [46], YouTube [41], Weibo [88], Reddit [74], and Flickr [20], the majority of previous studies have investigated emotions expressed after disasters on Twitter.

High quality data collection from social media can be a cumbersome task, especially for training a model with deep neural networks that require large datasets to perform well. Hence researchers have developed advanced algorithms to automatically identify, collect, and classify relevant data which also enable online monitoring of catastrophic events. For example, Ullah and his colleagues [86] employed rule-based systems and logistic regression to automatically collect tweets related to disasters and then classify them into subtypes like food, medical emergencies, and shelters. Since some of these algorithms require dictionaries of relevant words, disaster-related lexicons have been created and released to facilitate this process [65, 83].

Labeled data is another requirement of research to train models. These models can later suggest emotions for new unlabeled texts. There are currently several general-domain datasets widely used for emotion mining such as Semeval-2018 [55] but, in the context of disasters where negative emotions like anger, fear, and disgust might prevail, labeled domain-specific data can help with more fine-grained emotion recognition. Therefore, researchers have developed datasets that exclusively or predominantly contain disaster-related text with emotion labels [2]. A newer trend is creating multimodal datasets that contain both imagery and textual data [3]. However, to the best of our knowledge currently available multimodal datasets do not include emotion annotations.

**_Scope of the literature._** Another aspect of literature is the ever-expanding scope of countries, languages, and disasters covered by previous studies, although not necessarily in a systematic and informative way.

Disaster loss datasets (e.g., EM-DAT) categorize disasters into natural and technological categories, and there is empirical evidence suggesting that these categories provoke different patterns of affective responses [77]. While floods [11], earthquakes [13], hurricanes [90], and other natural disasters have been the focus of investigation, technological disasters have been rarely studied, with some exceptions like the Sewol Ferry Disaster in South Korea [41] and the West Texas Explosion [35].

Since there are cross-cultural differences in responses to disasters even among western countries [52], it is important to look at affective responses in different cultures. Disaster responses in individualistic cultures like the United States [38] and collectivist cultures like China and India [68], as well as in economically advanced countries like Germany [56] and upper-middle-income countries such as South Africa [43], have been studied. More importantly, this field has not restricted itself to English language; Hindi, Mandarin Chinese, Japanese, Korean, and many other languages have been subjects of study. So, the previous literature has provided good coverage of various cultures and languages.

***Emotion mining techniques.*** Similar to NLP which has evolved from simple rule-based systems to sophisticated deep neural networks [32] and now pre-trained large language models, emotion mining has experienced a lot of significant advancements over the years. From lexicon-based approaches [7], the field has progressed to deep neural networks [85], and researchers now utilize transformer-based pre-trained models like Generative Pre-training (GPT) and Bidirectional Encoder Representations from Transformers (BERT) [1, 37].

In the context of disasters, different ML algorithms have been used including Naïve Bayes and Support Vector Machines [12], Deep Learning models [15], Convolutional Neural Networks (CNN), and Long Short Term Memory [51]. Depending on the specifics of the methods such as datasets for training and test, vectorization methods, and evaluation techniques, high accuracy, for example above 80% [88] with CNN or even above 90% [61] with Bayesian Networks have been reported.

As ML algorithms continue to develop, so do NLP libraries that can automate and accelerate the emotion mining process. The Valence Aware Dictionary for Sentiment Reasoning (VADER; [24]) and TextBlob [47] are among the most popular libraries for sentiment analysis, and they have been used in the context of disasters. Although these general-purpose tools might not achieve the high accuracy of fine-tuned algorithms specifically developed for disasters [89], their ease of use has paved the way for those who lack the resources to develop their own tools. The development of automated tools for other languages than English, such as snowNLP for Chinese [31], has further expanded the reach and applicability of sentiment analysis in diverse linguistic and cultural contexts, making emotion mining more accessible globally.

*Applications*. If earlier work on emotion mining after disasters was basic and general and tend to be proof-of-concept demonstrating feasibility and confirming intuitions like prevalence of negative sentiment during or after disasters (e.g., [78]) the filed has shifted toward more nuanced, sophisticated analysis that can provide actionable insights. Innovative research such as significant event detection through monitoring nuanced changes in negative sentiment [6], investigation of emotional responses to different episodes of disasters [44], identifying the exact causes of anxiety and distress [59] and events that trigger changes in anger [81] has paved the way for applied work.

In sum, emotion mining after disasters has progressed from basic research to finding real-world applications. It is now possible to monitor affective responses to large scale catastrophes, detect areas and population at risk, and by providing mental health consultation reduce the risk of acute stress or post-traumatic stress disorders [30]. Considering the correlation between expressed emotions and actual damage, it is even possible to rapidly estimate damage after calamities by affective computing [11].

**From computer science to social and behavioral sciences**

Our review indicates that emotion mining in the context of disasters has significantly advanced over the years, benefiting greatly from the continuous improvements in natural language processing (NLP) techniques. Furthermore, the unique challenges and applications within disaster contexts have, in turn, helped advancements in NLP methodologies. With all the advancements, especially the advent of large language models and the use of generative AI in emotion mining [69] it is safe to assume we will see even more rapid development in the field. At the same time, some pervasive methodological issues persist which prevents the field from showing its full potential. Data-driven research on emotion mining during disasters has not yet informed theories, it has not fully benefited from existing theoretical frameworks and with a few exceptions it has ignored insights from social and behavioral sciences. In the next section, we discuss some of these issues.

**Drought of theories; Methodological issues in emotion mining during disasters**

With the availability of big data and data science methods, researchers can generate predictions by performing a series of data processing steps, often without incorporating theoretical insights into the subject matter [19]. This approach contrasts sharply with theory-driven research, in which researchers select a theoretical framework, develop a falsifiable hypothesis from it, and subsequently test this hypothesis by using inferential statistical methods. Actually, data-driven and theory-driven research have evolved along independent paths [50]. Even when using the same statistical methods, the goal of prediction in data-driven research and the goal of explanation in theory-driven research can lead to different approaches in variable selection, choice of methods, and interpretation of similar models [79]. Both approaches are necessary for advancement of science and have their own pros and cons but heavy reliance on one approach and ignoring the other one can be problematic [17].

Most previous studies on emotion mining in disasters have been conducted in a data-driven fashion, making it difficult to contextualized findings within a broader understating, is susceptible to biases which leads to skewed findings, and resulted in fragmented knowledge that is not easily synthesized into a comprehensive understanding of the impact of disasters. Below we describe three issues that are easily recognized and addressed in theory-based social sciences but are less discussed in data-driven emotion mining.

**Theories and models of emotions**. There are two broad categories of emotion theories: discrete and dimensional [21].

Discrete theories, starting with seminal work of Paul Ekman [18], "*An Argument for Basic Emotions*", posit that emotions have characteristics that make them different from other phenomena like moods and traits. These characteristics are rapid onset, short duration, unbidden occurrence, automatic appraisal, and coherence among responses. Moreover, each basic emotion has features that make it unique and different from other emotions: Distinct primate homologies, physiological correlates, universal antecedents, reported experiences, universal signals, behavioral tendencies, and appraisal tendency. As any other good scientific theory, basic emotion theory of Ekman, has offered a theoretical framework for incorporating novel findings as new empirical evidence emerges [36]. So, for example based on neuroimaging studies of emotion circuits or universal facial expressions, researchers might add to or remove from the original list of six basic emotions Ekman proposed (i.e., happiness, fear, anger, sadness, surprise, and disgust).

This is not how discrete emotions have been treated in emotion mining literature. For example, in their study of 2013 Ya'an earthquake of China, Yang and his colleagues [88] trained a CNN to identify six categories of emotions: positive, neutral, angry, anxious, fearful, and sad. It is not clear why disgust and surprise are dropped from their classification. But this arbitrary emotion classification does not have any scientific justification based on empirical research and theoretical framework of discrete emotions. Moreover, this kind of arbitrary taxonomy makes it difficult to compare, contrast and synthesize studies. For example, Munakata and Kobayashi [59], in their investigation of three major earthquake in Japan, considered two broad categories of positive and negative emotions then divide them into feelings of pleasure, enjoyment, anger, and sadness. So, while both studies mentioned here are emotion mining in the context of earthquakes in eastern cultures they cannot be easily synthesized into coherent literature.

In addition to the improper application of discrete emotion theories, another important category of emotion theories that has been relatively ignored in computer science is dimensional theories. Originally proposed by Russell [75], these theories represent emotions as continuous numerical values in multiple dimensions such as valence (pleasantness-unpleasantness) and arousal (excitement-calmness). Relative to resources for discrete sentiment analysis, resources for dimensional emotion mining are rare however they are attracting more attention in recent years [42]. Unlike polarity analysis (positive, neutral, and negative) which is more appropriate in the context of product reviews where researchers want to know what polar prevail, sentiment analysis based on dimensional theories can hugely improve current research by providing more fine-grained (real-valued) results in the context of disasters where negative emotions might dominate. These theories have been used in psychological studies of disasters [77] and can be a future direction for emotion mining of disasters.

Besides discreet and dimensional theories, there is another family of emotion models called appraisal theories. Originally introduced by Arnold [5] and further expanded by Lazarus [40], appraisal theories seek to explain how different emotions can arise from the same event, depending on the individual and the context. These theories propose that the differences in emotional responses are a result of the cognitive appraisals or evaluations that individuals make about the significance and implications of the event [58]. Thus, these theories consider how dimensions beyond valence and arousal, such as agency or goal relevance, may shape discrete emotions. While these theories have informed models of communication during crises and disasters [48], their potential to enhance the accuracy of sentiment analysis [73] has been underutilized in emotion mining.

**Cross-cultural frameworks**. Social and behavioral sciences in general, and psychology in particular struggle with the problem of overreliance on samples from Western, Educated, Industrial, Rich, and Democratic (WEIRD) cultures [60]. This is not the case in the emotion mining of disasters, which could actually be one of the strengths of the field. As previously noted, the existing literature encompasses a diverse range of cultures and languages. Surprisingly, despite this extensive coverage, due to the absence of cross-cultural comparisons, the literature does not allow for firm conclusions about the similarities and differences in cultural responses to disasters.

By providing theoretical frameworks such as individualism-collectivism or loose-tight cultures [27], cross-cultural psychology has enabled researchers to understand universal patterns as well as variations in responses to large-scale threats [66] in a systematic and informative way. While replicating emotion mining studies in a single, new culture or language is beneficial, a more comprehensive approach would involve examining how different cultures shape emotional responses to disasters [63]. By paying attention to these cultural dynamics, new doors might open, offering deeper insights into the interplay between culture and emotional responses in disaster scenarios. This direction enhances the robustness of emotion mining findings and enriches our understanding of the cultural context of emotional reactions to disasters.

**Misinformation and bias**. The ultimate goal of emotion mining is to understand the affective states of people exposed to or informed about disasters and use this information for disaster management [29]. However, on social media the true sentiment of the public is often obscured by misinformation and bias.

An extremely small but vocal minority of supersharers on Twitter produce and spread the majority of fake news and misinformation [9] and they can distort the meta-perception of most users [71]. Misinformation spread faster than true information in most cases and is more likely to be forwarded and spread by network users including in case of disasters [49]. On the one hand, in the context of crisis tweets charged by anxiety are more likely to be rumor [64]. On the other hand, tweets with negative sentiment, like anxiety, have higher virality [39]. Given this level of severity, researchers and practitioners should be cautious about the findings of previous studies that did not consider misinformation, and future research should not ignore it.

While methods exist to detect and correct misinformation about disasters [91], it is important to acknowledge the inherent biases in data collection on Twitter. When a catastrophic event happens some users tweet more frequently than others which skews the average sentiment [92]. Beside these individual differences, there might be group-level differences as well. People from affluent neighborhoods are more likely to use Twitter during disasters [92]. These social and geographical disparities that bias sampling have been long recognized and acknowledged and are typically discussed in social and behavioral sciences papers. However, they should be more emphasized in future data-driven research.

**Conclusion**

Among the four goals of science—description, explanation, prediction, and application—the social and behavioral sciences predominantly emphasize the former two, while computer science primarily focuses on the latter two. This divergence has led to theory-driven social sciences and data-driven computer science, each with its respective advantages and disadvantages. Data-driven research, unconstrained by theoretical frameworks, frequently employs innovative technologies across various contexts, often without prior hypotheses, but it may overlook valuable insights from previous studies. Conversely, theory-driven research systematically develops new lines of inquiry based on existing literature, yet it tends to be slow in adopting new, advanced technologies. Interdisciplinary collaboration can enhance the quality of research and publications for both computer and social scientists [57].

As reviewed above, data-driven emotion mining in the context of disasters has applied a wide range of up-to-date techniques to better understand human affective responses to catastrophes. Datasets are continuously growing larger, and new lexicons and libraries are being released regularly. More advanced and accurate emotion mining algorithms, especially by employing large language models, are being introduced to the field. However, creative but scattered journal articles that are difficult to synthesize may result in advancements in NLP but not necessarily in better disaster preparation and mitigation.

While it is helpful to accurately document positive and negative emotions that emerge as rescue operations commence or infrastructure is affected, it is even more beneficial to have a coherent body of literature that highlights universal trends as well as culture-specific patterns. Achieving this comprehensive understanding as quickly and accurately as possible requires advanced NLP techniques coupled with robust theoretical frameworks. Interdisciplinary research is essential to bridge these gaps and enhance our ability to respond effectively to disasters.

To advance the field of emotion mining in disaster contexts with an interdisciplinary approach, several specific recommendations should be considered. First, the arbitrary addition or removal of emotion dimensions or categories should be avoided; instead, emotion classifications should be grounded in well-established theories. Additionally, it is crucial to account for both individual and group-level differences, such as socioeconomic status or cultural background, to ensure a more nuanced understanding of emotional responses. While algorithmic data collection is a powerful tool, it does not automatically resolve issues related to misinformation and bias; therefore, researchers should actively check for or at least acknowledge possible biases in their studies. Moreover, the importance of replicating findings, a cornerstone in the social sciences, should not be overlooked in the pursuit of novelty within technical fields. Finally, the publication of null results is essential for the robustness of the literature, helping to mitigate the risk of publication bias and ensuring a more comprehensive understanding of emotion mining in disaster contexts.

When natural and technological disasters occur and emergency evacuation is underway, crowds of people demonstrate specific patterns of behavior [16]. Understanding these behaviors can be facilitated by more frequent collaboration between data-driven NLP and theory-driven social sciences. Only through such interdisciplinary efforts we can hope for more efficient, evidence-based disaster preparedness and mitigation.

**Author contributions**

The study was conceived and directed by the lead author. Two co-authors reviewed the NLP-related literature and drafted the discussion of technical details. The lead author took charge of writing the theoretical discussion. All authors provided critical feedback and contributed to the final manuscript.

**Statements and Declarations**

The authors declared no potential conflicts of interest with respect to the research, authorship, and/or

publication of this article

Ethical approval: Not Applicable

Data availability: Not Applicable

**Funding statement:**

The authors received no financial support for the research, authorship, and/or publication of this article.